\documentclass[sigconf]{acmart}

 \usepackage{listings}
\usepackage{xcolor}

\usepackage{balance}

\AtBeginDocument{%
  }

\copyrightyear{2025}
\acmYear{2025}
\setcopyright{cc}
\setcctype{by}
\acmConference[WWW Companion '25]{Companion Proceedings of the ACM Web Conference 2025}{April 28-May 2, 2025}{Sydney, NSW, Australia}
\acmBooktitle{Companion Proceedings of the ACM Web Conference 2025 (WWW Companion '25), April 28-May 2, 2025, Sydney, NSW, Australia}
\acmDOI{10.1145/3701716.3717539}
\acmISBN{979-8-4007-1331-6/2025/04}

\begin{document}

\title[Network-informed Prompt Engineering against Organized Astroturf Campaigns]{Network-informed Prompt Engineering against Organized Astroturf Campaigns under Extreme Class Imbalance}

\author{Nikos Kanakaris}
\authornote{Corresponding author: Nikos Kanakaris <kanakari@usc.edu>}
\email{kanakari@usc.edu}
\affiliation{%
  \institution{University of Southern California}
  \city{Los Angeles}
  \state{CA}
  \country{USA}
}

\author{Heng Ping}
\email{hping@usc.edu}
\affiliation{%
  \institution{University of Southern California}
  \city{Los Angeles}
  \state{CA}
  \country{USA}
}

\author{Xiongye Xiao}
\email{xiongye@usc.edu}
\affiliation{%
  \institution{University of Southern California}
  \city{Los Angeles}
  \state{CA}
  \country{USA}
}

\author{Nesreen K. Ahmed}
\email{nesahmed@cisco.com}
\affiliation{%
  \institution{Cisco AI Research}
  \city{Santa Clara}
  \state{CA}
  \country{USA}
}

\author{Luca Luceri}
\email{lluceri@isi.edu}
\affiliation{%
  \institution{University of Southern California}
  \city{Los Angeles}
  \state{CA}
  \country{USA}
}

\author{Emilio Ferrara}
\email{emiliofe@usc.edu}
\affiliation{%
  \institution{University of Southern California}
  \city{Los Angeles}
  \state{CA}
  \country{USA}
}

\author{Paul Bogdan}
\email{pbogdan@usc.edu}
\affiliation{%
  \institution{University of Southern California}
  \city{Los Angeles}
  \state{CA}
  \country{USA}
}


\renewcommand{\shortauthors}{Nikos Kanakaris et al.}

\begin{abstract}
Detecting organized political campaigns, commonly known as astroturf campaigns, is of paramount importance in fighting against disinformation on social media. Existing approaches for the identification of such organized actions employ techniques mostly from network science, graph machine learning and natural language processing. Their ultimate goal is to analyze the relationships and interactions (e.g. re-posting) among users and the textual similarities of their posts. Despite their effectiveness in recognizing astroturf campaigns, these methods face significant challenges, notably the class imbalance in available training datasets. To mitigate this issue, recent methods usually resort to data augmentation or increasing the number of positive samples, which may not always be feasible or sufficient in real-world settings. Following a different path, in this paper, we propose a novel framework for identifying astroturf campaigns based solely on large language models (LLMs), introducing a Balanced Retrieval-Augmented Generation (Balanced RAG) component. Our approach first gives both textual information concerning the posts (in our case tweets) and the user interactions of the social network as input to a language model. Then, through prompt engineering and the proposed Balanced RAG method, it effectively detects coordinated disinformation campaigns on $\mathbb{X}$ (Twitter). The proposed framework does not require any training or fine-tuning of the language model. Instead, by strategically harnessing the strengths of prompt engineering and Balanced RAG, it facilitates LLMs to overcome the effects of class imbalance and effectively identify coordinated political campaigns. The experimental results demonstrate that by incorporating the proposed prompt engineering and Balanced RAG methods, our framework outperforms the traditional graph-based baselines, achieving 2$\times$-3$\times$ improvements in terms of precision, recall and F1 scores.
\end{abstract}

\begin{teaserfigure}
\centering
\includegraphics[width=0.95\textwidth]{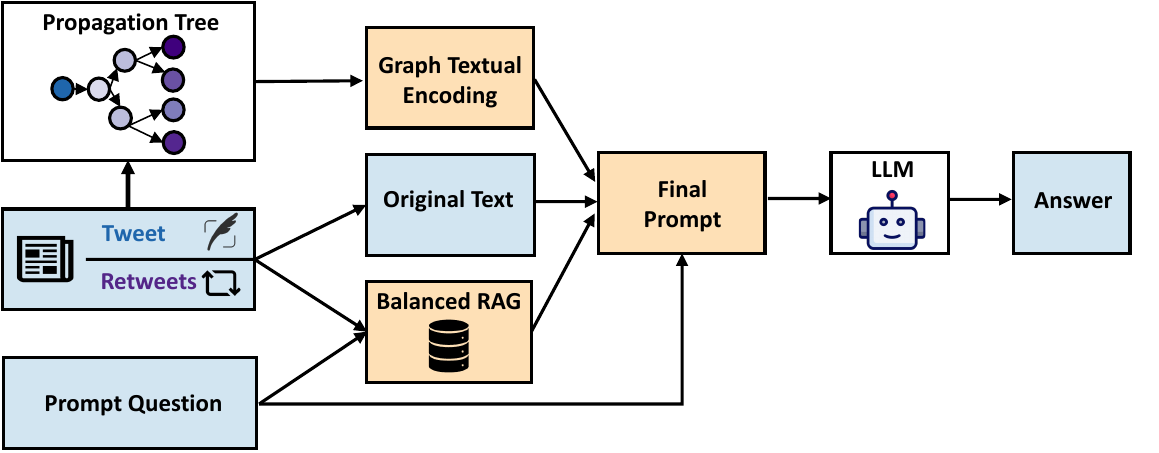}
\caption{Overview of the proposed framework. We first construct a propagation tree $G$ based on the posting and re-posting actions of a social network (e.g. $\mathbb{X}$/Twitter). Then, we employ graph prompting techniques to encode the produced propagation tree $G$ as a text, in an attempt to make information related to the structure of the social network available to an LLM. A similarity-based balanced RAG component provides several positive (fake news campaigns) and negative examples, which assist in better identifying astroturf campaigns. The text of the original tweet, the text-encoded graph and the examples from the RAG component are then combined with the initial prompt question to form the final prompt. Finally, an LLM predicts whether the given prompt concerns a coordinated disinformation campaign or not. 
Our framework is flexible and compatible with the most popular prompt engineering and RAG techniques. 
}
\label{fig:proposed-framework}
\end{teaserfigure}







\maketitle

\section{Introduction}
Social media have undoubtedly emerged as one of the most powerful mediums of information capable of influencing and even shaping public opinion~\cite{varlamis2022survey, fernandez2018online, tasnim2020impact}. They provide their users with an immediate, easy-to-use way of sharing and accessing rich content. Among other factors, their directness and convenience have definitely contributed to their wide adoption. Additionally, they facilitate and promote fruitful conversations among peers, allowing everyone to express their opinion freely~\cite{grinberg2019fake, bhattacharya2012sharing}.

Despite their widespread success, social media have also provided a fertile ground for malicious users to orchestrate and execute coordinated disinformation campaigns, commonly referred to as astroturf campaigns~\footnote{In this paper, the phrases `astroturf campaigns' and `disinformation campaigns' are used interchangeably.}~\cite{schoch2022coordination}. Such disinformation campaigns come to augment the negative effects of fake news by spreading persuasive but inaccurate content across a social network. They deliberately confuse and manipulate
people and public opinion, usually in an attempt to influence the outcome of a political decision, for instance during a presidential election~\cite{bovet2019influence, keller2017manipulate, keller2020political}. In most cases, the original coordinators of the campaigns camouflage their true identities by pretending to be trustworthy state representatives, towards promoting their political viewpoints~\cite{bunker2020you, freelon2020russian}. A common characteristic of such campaigns is their close connection to fake accounts. These fake accounts typically include fake, automated, state-sponsored accounts and social bots, which are created to artificially amplify the strength of a piece of information by artificial amplification. Their objective is to distribute content in a way resembling that of organic users~\cite{keller2017manipulate}. Adding to the complexity, it has been observed that specific groups of real users, such as conservative individuals and the elderly, are more susceptible to engaging with and re-sharing political disinformation~\cite{bovet2019influence, grinberg2019fake}.

More recently, with the advent of the generative artificial intelligence (GenAI) era, the proliferation and spread of fake news on social media have become even more problematic in modern society~\cite{ezzeddine2023exposing}. GenAI can further exacerbate the spread of disinformation by equipping malicious individuals with sophisticated tools to create compelling fake content, thus making it increasingly difficult for users to distinguish between real and fake news~\cite{ezzeddine2023exposing, loth2024blessingcursesurveyimpact}.

In an attempt to mitigate the effects of astroturf campaigns, existing approaches have primarily relied on graph mining and natural language processing (NLP) techniques.
Graph neural networks (GNNs) and large language models (LLMs) have emerged as the default technology to perform graph mining and NLP tasks, respectively.
The combination of LLMs and GNNs has also been proposed in the literature~\cite{michail2022detection, yi2020graph}.
GNNs extract structural and topological features associated with a propagation tree-shaped graph $G$ formed by the (re-)posting actions of the users, while LLMs and other NLP methods deal with the textual content of the posts.
The aforementioned propagation tree $G$ usually consists of a set of nodes $V$ denoting the posts and re-posts of the users and a set of edges $E$ denoting the re-posting actions (see Section~\ref{sec:definition} for an extensive problem definition).  Most of the existing approaches build on a GNN-based architecture wherein embeddings (vectors) from LLMs are attached to each node of the propagation tree. Briefly, the inclusion of textual embeddings enables GNNs to take into consideration both the content and the user interactions simultaneously.

Although recent approaches have generally been successful, there is still space for improvement. In particular, they display several shortcomings: (i) \textbf{Amounts of labeled data needed:} To begin with, they require large amounts of labeled data for training, which may not be available, mainly due to policy restrictions of the considered social media platforms~\cite{taylor2018mining}; (ii) \textbf{Class imbalance:} Even though the necessary data are available, given that the number of organic conversations is way larger than that of the fake news ones, the existing datasets demonstrate an extreme class imbalance; as a result, the available data critically inhibit the performance of the proposed methods~\cite{michail2022detection};
(iii) \textbf{Biased and not easily transferable:} The training process itself makes the models biased towards a specific dataset. Therefore, they are not easily transferable to new cases or environments; (iv) \textbf{Outdated, fostering hallucinations:} They are frequently trained on outdated data or data that promote LLM hallucinations~\cite{lewis2020retrieval}. (v) \textbf{Unsustainable or infeasible training:} Finally, training these methods, especially those including LLMs, is neither sustainable nor feasible due to their large number of trainable parameters or their closed-source code policies (e.g., OpenAI's GPT-4 model)~\cite{giray2023prompt, li2022survey}.

To address the above issues, in this paper, we propose a novel framework for identifying coordinated disinformation campaigns relying only on frozen LLMs. The proposed framework encodes both the structural and textual information of a propagation tree as a text, ready to be prompted to an LLM. Using prompt engineering and retrieval-augmented generation (RAG) allows LLMs to detect organized astroturf campaigns on social networks such as $\mathbb{X}$. Our framework does not involve any training or fine-tuning of the given LLM. In contrast, it capitalizes on the power of prompt engineering and RAG techniques and enables LLMs to overcome the effects of class imbalance, thereby effectively identifying coordinated political campaigns. Our framework is flexible and can be coupled with any prompt engineering and RAG technique (e.g. few-shot prompting)~\cite{fang2024multillmtextsummarization}. Furthermore, it is compatible with any decoder-only LLM.

To assess the robustness of our approach, we conducted several experiments and ablation studies on a dataset concerning the 2016 United States presidential election. The evaluation results reveal that the combination of prompt engineering and RAG yields better performance in terms of recall, F1 and ROC AUC scores against graph-based and other popular baselines.

\noindent\textbf{Motivation.} Our motivation is to design a framework that (i) does not require training and a lot of computation units to operate, (ii) handles class imbalance effectively without the need for ad-hoc feature engineering or data augmentation methods, (iii) is not affected by the frequent changes in the strategies of disinformation campaigns followed by the malicious users. 

\noindent\textbf{Contributions.} The contributions of this paper are as follows:
\begin{itemize}
    \item \textbf{Prompt engineering + RAG:} To the best of our knowledge, this is the first flexible framework that introduces the concepts of prompt engineering and RAG to the task of identifying coordinated disinformation campaigns.
    \item \textbf{Framework `class imbalance'-proof:} We propose a framework resilient to class imbalance.
    \item \textbf{Balanced RAG:} We also introduce `Balanced RAG', a new RAG method that assists in the class imbalance problem. 
    \item \textbf{Graphs are encoded as a text:} We propose a structure-aware prompting technique that encodes both the text and the graph of a coordinated astroturf campaign as a text, ready to be used from a frozen LLM.
    \item \textbf{Yields better performance} The proposed variants of the framework achieve a state-of-the-art performance of up to 42.4\%, 85.1\%, 56.6\% and 86.0\% in terms of precision, recall, F1 and ROC AUC scores, respectively.
\end{itemize}

\textbf{Research questions.} In addition, we answer the following research questions (Section~\ref{sec:discussion}):
\begin{itemize}
    \item \textbf{RQ1}: \textit{To what extent can frozen LLMs identify organized political disinformation campaigns?}: We show that frozen LLMs can effectively identify organized political disinformation campaigns, assuming that prompt engineering and a RAG component are provided.
    \item \textbf{RQ2}: \textit{Is graph information encoded as a text effective to detect organized political disinformation campaigns?}: We demonstrate that encoding graphs as text suffices to identify astroturf campaigns. 
    \item \textbf{RQ3}: \textit{Is it feasible to simply utilize prompt engineering and RAG techniques to identify political disinformation campaigns?}: We reveal that it is feasible to utilize only prompt engineering and RAG techniques to identify astroturf campaigns effectively using a frozen LLM.
    \item \textbf{RQ4}: \textit{Are prompt engineering and RAG reliable as far as the class imbalance problem is concerned?}: Our experimental results illustrate that prompt engineering and RAG techniques are capable of mitigating the effects of class imbalance.
    \item \textbf{RQ5}: \textit{Are GNNs really necessary to analyze graph-based data?}: We found that GNNs are not necessary to model graph-based data. Instead, strategically encoding graph data as text allows LLMs to improve their performance in terms of the task of identifying astroturf campaigns.
\end{itemize}

\section{Related work}
\label{appex:related}

\noindent \textbf{Prompt engineering.} Prompt engineering has emerged as a crucial tool in utilizing large language models for a variety of tasks, including disinformation detection. Zero-shot and few-shot prompting techniques, in particular, allow models to perform tasks without the need for extensive retraining or fine-tuning, which is crucial for adapting to rapidly changing disinformation campaigns \cite{giray2023prompt}. Role prompting and chain-of-thought prompting have been shown to improve the models' reasoning capabilities, enabling them to handle more complex tasks like the identification of astroturf campaigns \cite{luceri2024susceptibility}. Recent studies have demonstrated that sophisticated prompt selection strategies can significantly improve LLM performance in identifying coordinated campaigns under extreme class imbalance~\cite{fatemi2023talk}.

\noindent \textbf{Retrieval-augmented generation.}
Retrieval-augmented generation (RAG) techniques enhance the performance of LLMs by providing external, contextually relevant information during inference. RAG systems index documents and retrieve them based on a given query, mitigating the effects of outdated or incomplete knowledge in the LLM \cite{lewis2020retrieval}. In the context of disinformation detection, RAG allows models to retrieve propagation trees that are similar to the given input, improving the detection accuracy of coordinated fake news campaigns. The integration of RAG with prompt engineering has proven to be a potent strategy for tackling class imbalance in disinformation datasets~\cite{varlamis2022survey}.

\noindent\textbf{Graph neural networks and large language models.}
Graph-based approaches have traditionally been used to detect disinformation campaigns by analyzing the relationships between users and posts on social media platforms. Graph neural networks (GNNs), such as those employed by~\cite{michail2022detection} and \cite{mitra2024usegraphneuralnetworks}, leverage the structure of social networks to detect coordinated behavior. However, recent developments have shown that textual encoding of graph structures into prompts for LLMs, without the need for GNNs, can be equally effective. This approach is particularly useful in scenarios where training a GNN is computationally expensive or infeasible due to the dynamic nature of social media disinformation campaigns \cite{monti2019fake}.

\noindent\textbf{Detecting coordinated disinformation campaigns.}
The detection of coordinated disinformation campaigns, commonly referred to as astroturfing, has been a growing area of research. Traditional methods focus on using social network analysis and natural language processing (NLP) to identify fake accounts and bots that spread misinformation~\cite{keller2020political}. Recent works have integrated both the structural and textual information of social media posts into a unified framework that employs prompt engineering and retrieval-augmented generation, allowing frozen LLMs to identify coordinated campaigns without the need for additional training or fine-tuning~\cite{bhattacharya2012sharing}. This approach has been shown to outperform graph-based baselines, particularly in terms of recall and F1 scores, under extreme class imbalance settings~\cite{luceri2024susceptibility}.

\section{Proposed framework}
\label{sec:framework}

\subsection{Framework overview}

In this section, we present our framework. Briefly, it is composed of four phases as illustrated in Figure~\ref{fig:proposed-framework}. Initially, we generate a propagation tree based on the posting and re-posting~\footnote{Since our dataset comes from $\mathbb{X}$, in the context of this paper, posts and re-posts correspond to tweets and re-tweets, respectively.} actions of a social network. Then, using graph prompting techniques, we encode the generated propagation tree as a text, aiming to make graph information related to the social network available to a given LLM. In the case of few-shot prompting, a RAG component provides the overall pipeline with several labeled examples, potentially assisting in the final task. The text of the original tweet, the text-encoded graph and the examples from the RAG component are then combined with the initial prompt question to form the final prompt. Finally, an LLM predicts whether the given prompt concerns a coordinated disinformation campaign or not. We note that all the necessary background is presented in Section~\ref{appex:related}.


\subsection{Problem definition}
\label{sec:definition}
In this section, we define important notation and formulate our main problem.

\begin{definition}
\label{def:user}
\textbf{(Users and tweets)}.
Let $U = \{u_1, u_2, \dots, u_{|U|}\}$ be the set of users of a social network. Let also $T = \{t_1, t_2, \dots, t_{|T|}\}$ be the set of the textual documents of tweets and re-tweets. Each user $u$ is the author of a subset of tweets and re-tweets $T_{u} \subseteq T$. Furthermore, each user $u$ has a number of followers $k \in \mathbb{N}^+$.
\end{definition}

\begin{definition}
\label{def:tree}
\textbf{(Propagation tree)}.
We denote a propagation tree as $G = (V, E)$. Essentially, $G$ is a tree-shape graph consisting of a set of nodes $V = \{v_1, v_2, \dots, v_{|V|}\}$ and a set of edges $E \subseteq V \times V$. Each node $v \in V$ corresponds to a tweet or re-tweet and is related to a user $u \in U$ and a textual document $t \in T$. Each edge $e=(v_i, v_j) \in E$ denotes a re-tweet action, where re-tweet $v_j$ re-tweeted (re-)tweet $v_i$. In our case, the root node $v_1$ of a propagation tree $G$ is always a tweet. Similarly, each node $v_i, i \geq 2$ is a re-tweet. We also compute the time (delay) difference $d_j \in \mathbb{R}^+$ between two tweets $v_i$ and $v_j$. We note that the delay $d_1$ of the original tweet $v_1$ is $0$. To sum up, a propagation tree represents re-posting (re-tweeting) actions among a tweet $v_1$ and a set of re-tweets $\{v_2, \dots, v_{|V|}\}$.

\end{definition}

\begin{definition}
\label{def:problem}
\textbf{(Main problem)}. We focus on the problem of identifying coordinated disinformation campaigns on social media (e.g. $\mathbb{X}$) under extreme class imbalance using only a frozen LLM $f$ for inference as well as prompt engineering and RAG techniques. Given a propagation tree $G = (V, E)$ and the text $t_1$ of the original tweet $v_1$, our goal is to predict whether the propagation tree $G$ is an organized disinformation campaign or not.
\end{definition}

\subsection{Propagation tree construction}
\label{sec:propagation}
The first phase of the proposed framework is to construct the propagation tree $G$. Depending on the social network under consideration this step can be omitted. Here, this step is critical, given that the information provided by $\mathbb{X}$ is a star graph where all the re-tweets $\{v_2, \dots, v_{|V|}\}$ are linked to the original tweet $v_1$. Therefore, to predict the re-tweeting actions $E$ of a propagation tree $G$, we adopt the method proposed in~\cite{monti2019fake, michail2022detection}. We first sort the set of (re-)tweets $V$ by time in ascending order. Then, for each re-tweet $v_i \in V$, where $2 \leq i \leq |V|$, we add an edge $e=(v_j, v_i) \in E$ iff:
\begin{itemize}
    \item \textbf{Condition 1:} The text $t_i$ of the re-tweet $v_i$ includes an {\sc RT} mention to a specific user $u \in U$ who is the author of a retweet $v_j \in \{ v_1, \ldots, v_{i-1} \}$.
    \item \textbf{Condition 2:} The author $u_i$ of (re-)tweet $v_i$ follows the author $u_j$ of a tweet $v_j \in \{ v_1, \ldots, v_{i-1} \}$. If multiple such users exist, we pick the most popular, i.e. the one with the most followers $k$.
\end{itemize}

\noindent Finally, if none of the above conditions applies, we randomly select a (re-)tweet $v_j \in \{ v_1, \ldots, v_{i-1} \}$, using the number of followers $k$ of each $u$ as a selection criterion. This process follows a discrete weighted probability distribution, where the probability of selecting tweet $v_j$ is proportional to the number of followers $k_j$ of user $u_j$ who posted that tweet. Formally, the probability $P \in [0, 1]$ of selecting tweet $v_j \in \{v_1, v_2, \dots, v_{i-1}\}$ is given by
\begin{equation}
    P(v_j) = \frac{k_j}{\sum^{i-1}_{m=1}k_m}
\end{equation}

\subsection{Prompt engineering}
We now describe the prompt engineering phase of the framework. This phase aims to encode both the structural and textual information of a propagation tree $G$ as a text, in a way that the overall performance of a given LLM $f$ is generally improved. 

\noindent\textbf{Graph encoding.} As opposed to existing approaches that employ GNNs to encode the topological characteristics of a propagation tree $G$, we decide to follow a new paradigm~\cite{fatemi2023talk, perozzi2024let} where both the graph and text are given as a textual prompt to an LLM. To that end, we introduce the graph encoding function $g: G \rightarrow W$, where $W = \{w_1, w_2, \dots, w_{|W|}\}$ is the set of the available tokens (words). The graph encoding function $g$ takes as input a propagation tree $G$ and converts it into a set of words $W_g \subseteq W$.
Following the encoding guidelines presented in~\cite{fatemi2023talk}, we found that the majority of the LLMs demonstrate an improved performance by simply encoding the re-tweet actions $E$ of a propagation tree $G$ as a text as follows: \textit{"(2->1), (3->1), (4->3)"}. In the previous example, $E = \{(1, 2), (1, 3), (3, 4)\}$ and $|V|=4$. We note that other graph encoding functions can also be used based on the problem and the nature of the given graph $G$, including `adjacency', `incident' and `friendship' encoding. We refer to~\cite{fatemi2023talk} for an extensive list of graph encoding functions for LLMs.

\noindent\textbf{Prompting techniques.} Most of the available prompting techniques are compatible with our framework, including zero-shot, few-shot, chain-of-thought (CoT) and role prompting. Considering zero-shot prompting, our approach combines the textual document $t_1$ of the original tweet $v_1$ and the text-encoded version of the structural information $E$ of a propagation tree $G$ to create the final prompt. As a result, zero-shot prompting relies only on the pre-trained knowledge of the LLM to generate a response.

Although zero-shot prompting can be a sufficient technique, few-shot prompting is much preferable when labeled examples of positive (fake news) and negative propagation trees are available. These examples can also be incorporated into the final prompt and therefore allow the model to gain some knowledge from them, in an attempt to improve its performance. The prompted examples can be selected either at random or using a retrieval method. We found that utilizing a sophisticated method to strategically select the prompting examples increases the performance of an LLM significantly. We further discuss the selection of examples in Section~\ref{sec:rag}.

Chain-of-thought (CoT) prompting enables LLMs to break down complex problems into more intuitive steps. CoT prompting has shown strong performance in tasks requiring calculations. In our case, CoT prompting allows LLMs to effectively analyze the structural information $E$ of a propagation tree $G$.

Finally, a combination of prompting techniques can also be followed. We found that mixing role prompting, CoT and few-shot prompting yields the best results~\cite{cheng2024structureguidedpromptinstructing}. A list of prompt examples considering the problem of detecting astroturf campaigns is available in Appendix~\ref{appex:prompts}. We refer to Section~\ref{appex:related} for a more thorough explanation of prompt engineering techniques.

\subsection{Balanced Retrieval-Augmented Generation (Balanced RAG)}
\label{sec:rag}
Here we introduce the Balanced Retrieval-Augmented Generation (Balanced RAG) component of our proposed framework, which is pivotal in addressing the challenge of extreme class imbalance in datasets, particularly when the number of positive samples significantly differs from the number of negative samples.

\noindent\textbf{Motivation and rationale.} Class imbalance is a prevalent issue in machine learning tasks, especially in scenarios like detecting coordinated disinformation campaigns where positive instances (e.g., fake news propagation trees) are scarce compared to negative instances (e.g., organic propagation trees). Traditional models trained on such imbalanced datasets tend to be biased towards the majority class, resulting in poor detection performance on the minority class. Balanced RAG aims to mitigate this issue by constructing pairs of samples that are highly similar in content and structure but have \emph{opposite labels}. By presenting these contrasting yet similar examples to the LLM, we enable it to focus on subtle differences between classes, enhancing its discriminative capabilities and effectively addressing class imbalance.


\noindent \textbf{Basic overview.} The Balanced RAG component operates by first retrieving the top \( n \) samples from the training dataset \( \mathcal{D} \) that are most similar to a given query propagation tree \( G' \). For each retrieved sample \( G \), regardless of its label (positive or negative), the method finds a sample with the \emph{opposite label} that has the highest similarity to \( G \). This process results in a set of balanced pairs \( P = \{ (G_1, G_{1,\text{opposite}}), (G_2, G_{2,\text{opposite}}), \dots, (G_n, G_{n,\text{opposite}}) \} \), where each pair consists of two highly similar samples with contrasting labels.

These balanced pairs are then incorporated into the few-shot prompting approach used to query the LLM. By providing the LLM with representative examples that are both similar to the query and equally cover both classes, we enhance the model's ability to distinguish between positive and negative instances. This approach enables the LLM to learn the critical differences that define each class, thereby improving its ability to correctly classify new, unseen propagation trees~\cite{au2025personalizedgraphbasedretrievallarge}.

\noindent\textbf{Indexing.} To facilitate efficient retrieval, each propagation tree \( G \) in the training dataset \( \mathcal{D} \) is represented as an embedding vector \( \mathbf{e} \in \mathbb{R}^d \) that captures both structural and textual features of the propagation tree. The indexing function is formally defined as:
\begin{equation}
\text{index}: G \rightarrow \mathbf{e} \in \mathbb{R}^d.
\end{equation}
These embeddings \( \mathbf{e} \) can be obtained using pre-trained LLMs that encode textual information, or simple vectorization methods like bag-of-words. Structural features of the propagation trees, such as node degrees or graph motifs, can be concatenated to the textual embeddings to enrich the representation. The resulting embeddings are stored in an index structure to enable efficient similarity computations during the retrieval phase.


\noindent\textbf{Retrieval process.} The retrieval process is a critical component of the Balanced RAG method, designed to select relevant and informative examples that aid the LLM in making accurate predictions, especially in the context of extreme class imbalance. It involves the following detailed steps:

\begin{itemize}
    \item \textbf{Retrieve top \( n \) similar samples to \( G' \).} Given a query propagation tree \( G' \), our objective is to find the most similar samples in the training dataset \( \mathcal{D} \) to provide contextually relevant examples to the LLM. We compute the similarity between \( G' \) and each sample \( G \) in \( \mathcal{D} \). The similarity score between two propagation trees \( G \) and \( G' \) is calculated as:
    \begin{equation}
    \text{similarity}(G, G') = \frac{\text{index}(G) \cdot \text{index}(G')}{\| \text{index}(G) \| \, \| \text{index}(G') \|},
    \end{equation}
    where \( \text{index}(G) \cdot \text{index}(G') \) denotes the dot product of the embedding vectors, and \( \| \cdot \| \) denotes the Euclidean norm.

    We then select the top \( n \) samples with the highest similarity scores to form the set \( D = \{ G_1, G_2, \dots, G_n \} \), where each \( G_i \in D \) is ranked based on its similarity to the query \( G' \).

    \item \textbf{For each \( G \in D \), find the most similar sample with opposite label.} For each sample \( G \in D \), we aim to find a sample \( G_{\text{opposite}} \) in the training dataset \( \mathcal{D} \) that has the opposite label to \( G \) and maximizes the similarity to \( G \):
    \begin{equation}
    G_{\text{opposite}} = \operatorname*{arg\,max}_{G' \in \mathcal{D}_{\text{opposite}}} \text{similarity}(G, G'),
    \end{equation}
    where \( \mathcal{D}_{\text{opposite}} = \{ G' \in \mathcal{D} \mid \text{label}(G') \ne \text{label}(G) \} \) is the subset of \( \mathcal{D} \) containing samples with labels opposite to that of \( G \).

    \item \textbf{Construct balanced pairs.} After identifying \( G_{\text{opposite}} \) for each \( G \in D \), we construct the set of balanced pairs:
    \begin{equation}
    P = \{ (G_1, G_{1,\text{opposite}}), (G_2, G_{2,\text{opposite}}), \dots, (G_n, G_{n,\text{opposite}}) \},
    \end{equation}
    where each pair \( (G_i, G_{i,\text{opposite}}) \) consists of two samples that are highly similar in terms of their embeddings but belong to opposite classes.
\end{itemize}

By incorporating both the most similar samples and their most similar counterparts from the opposite class, we not only mitigate the class imbalance problem but also expose the model to nuanced differences between structurally similar but differently labeled instances.


\subsection{Detecting astroturf campaigns}
In the last phase,  a frozen LLM $f$ predicts whether the considered propagation tree $G'$ concerns a coordinated disinformation campaign or not. Here, any LLM following a decoder-only generative architecture can be used. Formally, let $f: W \rightarrow W$ be a frozen LLM with non-trainable parameters that gets as input a final prompt $x \in W$ and returns an answer $a \in W$. Let also $q \in W$, $W_g = g(G) \in W$, $t_1 \in T$, $D \subseteq \mathcal{D}$ be a prompt question, the graph textual encoding of propagation tree $G$, the textual document of original tweet $v_1 \in V$, and the propagation trees examples obtained either randomly or through the RAG component, respectively.
We then combine $q, W_g, t_1$ and $D$ to create the final prompt $x$. We aim to find the best prompt combination such that the average performance of the model $f$ over the test set $S$ of the dataset is maximized with respect to a predefined metric:

\begin{equation}
\max_x \frac{1}{|S|} \sum_{i=1}^{|S|} score_f(g_i, W_{g_i}, t_{1_i}, a_i)
\end{equation}

where $a_i \in W$ represents the correct answer for sample $i$, and score$_f$ denotes the model $f$'s score on each test instance. Since we are primarily concerned with identifying positive samples (fake news campaigns), recall and $F_1$ are appropriate candidates for the scoring metric.  Depending on the selected prompting technique, we ask the model to respond either with a simple \textit{"Yes"} or \textit{"No"}, or with the answer followed by its chain of thought.

\section{Experiments}
\label{sec:experiments}

\subsection{Dataset}
    \label{subsec:dataset}

Our dataset is part of the FakeNewsNet dataset and concerns the 2016 United States presidential election. Each sample consists of a post from $\mathbb{X}$ and is linked to fake or real news based on Politifact. Since $\mathbb{X}$ does not provide information about the original re-tweeting actions, we predict the propagation tress $\mathcal{G}$ as described in Section~\ref{sec:propagation}. The final dataset includes $10228$ propagation trees, where $9.7\%$ of them are considered as coordinated political fake news campaigns. It involves $504,085$ unique users, where $7.8\%$ of them are malicious or susceptible to astroturf campaigns.
The high degree of class imbalance makes this dataset suitable for evaluating the proposed framework.
Table~\ref{tab:descriptiveStatistics} provides further statistics of the produced dataset.
    
\begin{table}[htbp]
    \centering
    \caption{Statistics of the dataset.}
    \label{tab:descriptiveStatistics}       
    \begin{tabular}{cc}
    \toprule
    {\bf Feature} & {\bf Value} \\
    \midrule    
    Number of users $|U|$ & $504,085$  \\
    Number of tweets $|V|$ & $65,000$ \\
    Number of propagation trees $|\mathcal{G}|$ & $10,228$ \\
    Number of positive (fake news) & $993$ \\
    Percentage of positive (fake news) samples & $9.7\%$ \\
    \bottomrule
    \end{tabular}
    \end{table}

\subsection{Setup}
We perform the necessary LLM inference of our experiments using the Ollama library. Our baseline models are implemented using the PyTorch Geometric library.

\noindent\textbf{Evaluation metrics.} Since it is important to identify the positive samples of a dataset (i.e., fake news campaigns), we selected precision, recall, F1 score and ROC AUC scores as our main evaluation metrics.

\subsection{Baselines and framework variants}
\noindent \textbf{Baselines.} To test the performance of the proposed framework, we benchmark several of its variants against a list of common baseline methods. The baseline models may differ from our variant models with regard to the architecture they follow or the type of features they consider. Additionally, they may ignore components of the proposed framework or
implement them differently. The main goal of the baselines is to evaluate our framework as far as the task of identification of coordinated campaigns is concerned. Finally, we note that the baseline models follow a GNN-based architecture, as proposed in~\cite{michail2022detection, varlamis2022survey}.


\noindent \textbf{Framework variants.} We present the most important framework variants used to evaluate our framework. In all cases, we use the Llama 3.1 70B model. Our framework variants differ from one another in terms of the selected prompt engineering technique and the design of the RAG component. The first variant namely, \textit{Llama 70B (0-shot)}, does not incorporate a RAG component and uses zero-shot prompting. The second variant, \textit{Llama 70B (6-shot)}, does not include a RAG component but adapts a few-shot prompting architecture, where 6 propagation trees are given as examples to the Llama model. The examples are randomly chosen. The third framework variant, \textit{Llama 70B + RAG}, employs few-shot prompting along with RAG. These six examples are the most similar to the propagation tree under investigation. The final framework variant, \textit{Llama 70B + Balanced RAG}, employs few-shot prompting and RAG. We strategically retrieve the three most similar positive (organic propagation trees) and the three most similar three negative (astroturf propagation trees) examples. In that way, we provide Llama with a balanced version of propagation tree examples. We refer to this variant as `balanced RAG' (see Section~\ref{sec:rag}). A detailed list of framework variants is also presented in Section~\ref{sec:ablation} for the ablation study.

\subsection{Evaluation results}
We compared our proposed framework against traditional graph-based baseline models, including GAT, GCN, GraphSAGE, and GraphConv, to evaluate its effectiveness in detecting coordinated disinformation campaigns under extreme class imbalance conditions. As detailed in Table~\ref{tab:evaluation}, our framework significantly outperforms these baselines across all evaluation metrics— precision, recall, F1, and ROC AUC. Specifically, the incorporation of the Balanced Retrieval-Augmented Generation (Balanced RAG) component further enhances the accuracy of our approach. For instance, our framework achieves the best performance with a recall of 0.8507, represent a 1.5$\times$-3$\times$ improvement over the baseline models, which suffer from the negative effects of class imbalance, exhibiting low scores across all evaluation metrics.

Considering the above remarks, we conclude that the combination of network-informed prompt engineering and Balanced RAG can definitely assist a frozen LLM in improving its performance with respect to the task of identifying coordinated fake news campaigns under extreme class imbalance.

    \begin{table}[htbp]
    \centering
    \caption{Main evaluation results of different baseline models and framework variants. The best results for each metric are highlighted in bold. The horizontal line distinguishes the baseline models from the framework variants. The best results for each metric are highlighted in bold.}
    \label{tab:evaluation}       
    \resizebox{\columnwidth}{!}{
    \begin{tabular}{lcccc}
    \toprule
    {\bf Method} & {\bf Precision} & {\bf Recall} & {\bf F1} & {\bf ROC AUC} \\
    \midrule    

    GAT & $0.1170$ & $0.5500$ & $0.1930$ & $0.5360$   \\
    GCN & $0.1820$ & $0.2500$ & $0.2110$ & $0.6070$   \\
    GraphSAGE & $0.1560$ & $0.4452$ & $0.2310$ & $0.6150$   \\
    GraphConv & $0.1311$ & $0.4550$ & $0.2040$ &  $0.5660$   \\\hline
    
    Llama 70B (0-shot) & $0.4072$ & $0.5348$ & $0.4624$ & $0.7236$  \\
    Llama 70B (6-shot) & $0.3048$ & $0.6816$ & $0.4212$  & $0.7533$\\  
    Llama 70B + RAG& $0.3707$ & $0.8308$ & $0.5127$ & $0.8360$ \\  
    Llama 70B + Balanced RAG& $\textbf{0.4238}$ & $\textbf{0.8507}$ & $\textbf{0.5658}$ & $\textbf{0.8602}$ \\  
    \bottomrule
    \end{tabular}
    }
    \end{table}

\subsection{Hyperparameter tuning}
We explore the impact of different hyperparameters on the performance of the proposed framework. To speed up the process, most of the experimentation on hyperparameter tuning was conducted using the Llama 3.1 8B model, unless otherwise stated.

\noindent\textbf{Impact of the temperature value}. We assess the impact of temperature value on the overall performance of the framework. The temperature value in LLMs controls the randomness of the model's predictions. It adjusts how confident or creative an LLM is regarding its responses. As shown in Table~\ref{tab:temperature}, the selected LLM model yields better results with a temperature of 0. This suggests that a model should be more deterministic.

\begin{table}[htbp]
    \centering
    \caption{Evaluation results of a framework variant with respect to different temperature values ranging from 0 to 1. The best results for each metric are highlighted in bold. Llama 3.1 8B was used in these experiments.}
    \label{tab:temperature_evaluation}       
    \resizebox{\columnwidth}{!}{
    \begin{tabular}{ccccc}
    \toprule
    {\bf Temperature} & {\bf Precision} & {\bf Recall} & {\bf F1} & {\bf ROC AUC}  \\
    \midrule    
    0.0 & \textbf{0.1303} & \textbf{0.9279} & \textbf{0.2286} & \textbf{0.6153}  \\
    0.2 & 0.1288 & 0.9179 & 0.2259 & 0.6094  \\
    0.4 & 0.1285 & 0.9204 & 0.2255 & 0.6087  \\
    0.6 & 0.1277 & 0.9005 & 0.2237 & 0.6040  \\
    0.8 & 0.1297 & 0.9129 & 0.2272 & 0.6116  \\
    1.0 & 0.1272 & 0.9054 & 0.2230 & 0.6079  \\
    \bottomrule
    
    \end{tabular}
    }
    \label{tab:temperature}
\end{table}


\noindent\textbf{Impact of the number of prompting examples.} We investigate how the number of prompting examples affects the overall performance of the framework. We mention that the number of prompting examples (propagation trees) may correspond to a different prompting technique. For instance, a number of prompting examples $n = 0$ requires 0-shot prompting, while $n = 5$ is compatible with few-shot prompting.
Table~\ref{tab:num_shots_evaluation_flipped} summarizes the results of our experiments. As shown, a value $n > 1$ performs better. This suggests that few-shot prompting techniques should generally be preferred.

\begin{table}[htbp]
    \centering
    \caption{Evaluation results of a framework variant with respect to different numbers of prompting examples ranging from 1 to 8. The best results for each metric are highlighted in bold. Llama 3.1 8B was used in these experiments.}
    \resizebox{\columnwidth}{!}{
    \begin{tabular}{ccccc}
    \toprule
    \textbf{Number} & \textbf{Precision} & \textbf{Recall} & \textbf{F1} & \textbf{ROC AUC} \\
    \midrule    
    1 & 0.1194 & 0.9030 & 0.2109 & 0.5766 \\
    2 & 0.1259 & 0.9154 & 0.2214 & 0.6000 \\
    3 & 0.1299 & \textbf{0.9303} & 0.2280 & 0.6143  \\
    4 & 0.1367 & 0.9154 & 0.2379 & 0.6322 \\
    5 & \textbf{0.1428} & 0.9005 & 0.2465 & 0.6459 \\
    6 & 0.1379 & 0.9080 & 0.2395 & 0.6345 \\
    7 & 0.1414 & 0.9055 & 0.2446 & 0.6432  \\
    8 & 0.1426 & 0.9204 & \textbf{0.2469} & \textbf{0.6486}  \\
    \bottomrule
    \end{tabular}
    }
    \label{tab:num_shots_evaluation_flipped} 
\end{table}

\begin{figure}[t]
    \centering
    \includegraphics[width=\columnwidth]{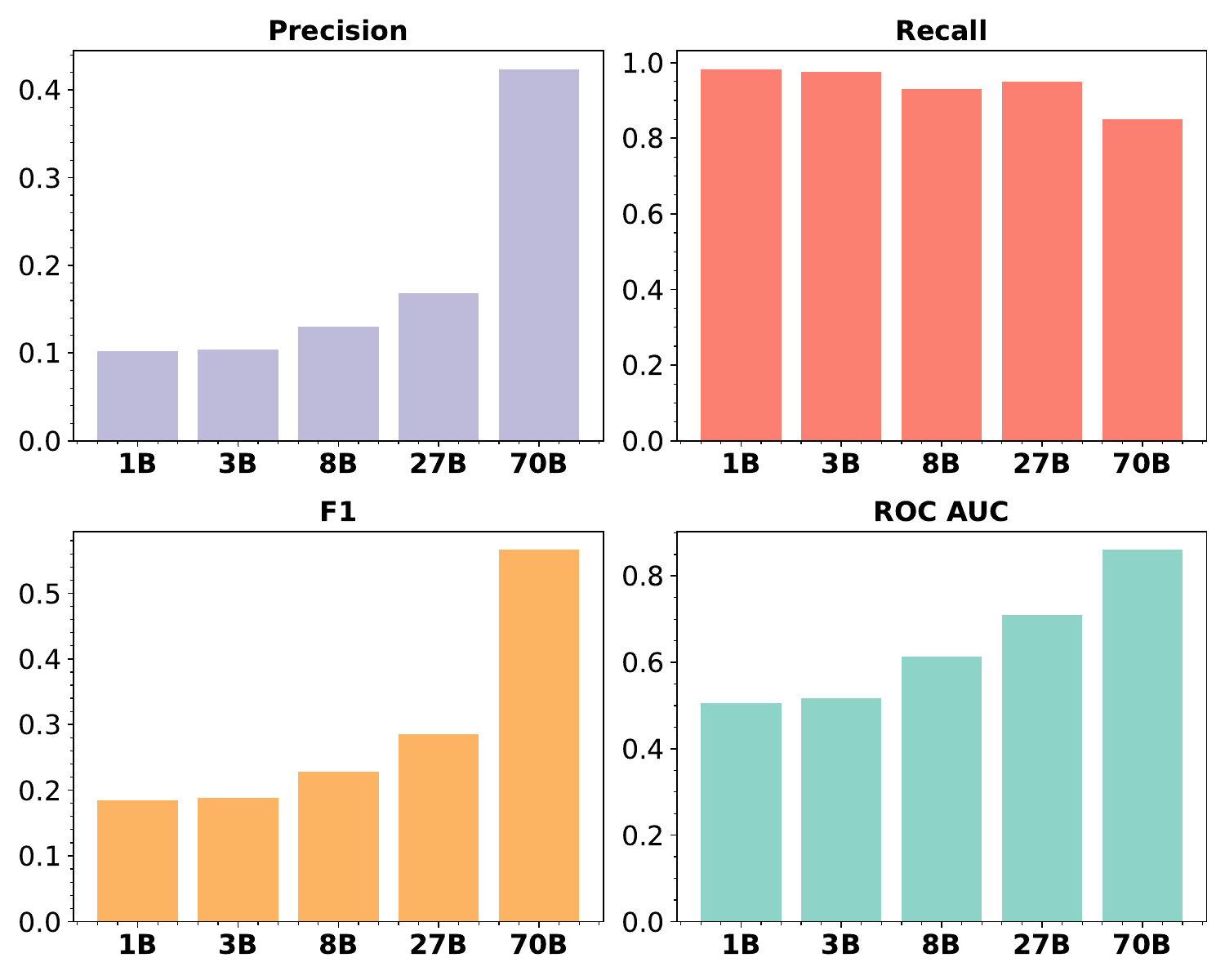}
    \caption{Comparison of different metrics (precision, recall, F1 score, and ROC AUC) across various (Llama) model sizes. The graph illustrates how each metric changes as the number of model parameters increases.}
    \label{fig:param}
\end{figure}

\noindent\textbf{Impact of the size of the LLM model.} We investigate how the different model sizes affect the overall performance of the framework. We run several experiments with Llama models of different parameter sizes, ranging from 1B to 70B parameters. Figure~\ref{fig:param} showcases the performance of each Llama model in terms of the precision, recall, F1 and ROC AUC scores. We conclude that models with more parameters perform better compared to smaller ones.


\section{Ablation study}
\label{sec:ablation}
We now investigate the contribution of each component to the performance of the proposed framework. We run the experiments for the ablation study with Llama 3.1 70B.







\subsection{Impact of prompt engineering and Balanced RAG}
To assess the impact of the RAG component on the overall performance of the framework, we run experiments with different RAG approaches. We also conduct experiments with numerous prompt engineering techniques to evaluate their overall impact. Results are shown in Table~\ref{tab:abl_prompt}. We describe each ablation study in detail below. Overall, the results indicate that combining the proposed balanced RAG method (i.e. retrieving positive and negative examples) with 6-shot prompting yields the best results.

\noindent\textbf{No RAG (w/o RAG).} We do not use any RAG component. Zero-shot prompting is also considered an approach that inherently does not make use of RAG. In the case of 1-shot and few-shot prompting, we randomly choose the propagation tree examples from the training set of the dataset. As shown in Table~\ref{tab:abl_prompt}, methods without a RAG component demonstrate the lowest results.

\noindent\textbf{RAG with one-shot (RAG (1-shot)).} We use a RAG component by selecting the most similar propagation tree example from the training set of the dataset, regardless of its label. We follow a one-shot prompting strategy. The inclusion of a RAG component and 1-shot prompting augments the performance of our framework.

\noindent\textbf{RAG with few-shot (RAG (X-shot)).} We use a RAG component by selecting $X$ propagation tree examples from the training set of the dataset, regardless of their labels. We perform few-shot prompting. The proposed combination results in better performance.

\noindent\textbf{RAG with similarity-based selection (Balanced RAG).} We perform few-shot prompting classification where we choose a predefined number of the most similar positive and similar negative samples of the training set of the dataset. We refer to this RAG method as `Balanced RAG'. As shown in Table~\ref{tab:abl_prompt}, this is the best combination of a RAG method and a prompt engineering technique. 

\begin{table}[htbp]
    \centering
    \caption{Evaluation results of the different combinations of RAG and prompt engineering approaches. The best results for each metric are highlighted in bold. The best results for each metric are highlighted in bold. Llama 3.1 70B was used in these experiments.}
    \label{tab:rag_evaluation}       
    \resizebox{\columnwidth}{!}{
    \begin{tabular}{lcccc}
    \toprule
    {\bf RAG Approach} & {\bf Precision}  & {\bf Recall} & {\bf F1} & {\bf ROC AUC}  \\
    \midrule    
    {\bf 0-shot} & 0.4072 & 0.5348 & 0.4624 & 0.7236  \\
    {\bf w/o RAG (1-shot)} & 0.3380 & 0.6667 & 0.4485 & 0.7589  \\
    {\bf w/o RAG (6-shot)} & 0.3048 & 0.6816 & 0.4212 & 0.7533  \\
    {\bf RAG (1-shot)} & 0.2880 & 0.7413 & 0.4156 & 0.7678  \\
    {\bf RAG (6-shot)} & 0.3707 & 0.8308 & 0.5127 & 0.8360  \\
    {\bf Balanced RAG} & \textbf{0.4238} & \textbf{0.8507} & \textbf{0.5658} & \textbf{0.8602}  \\
    \bottomrule
    \end{tabular}
    }
    \label{tab:abl_prompt}
\end{table}

\subsection{Impact of the encoded information}
We now investigate the influence of the encoded information on the performance of the proposed framework. Table~\ref{tab:encoded_evaluation} summarizes the overall results. As illustrated, providing an LLM with both the text of the tweets and the text-encoded propagation tree produces the best outcome for the task of identifying astroturf campaigns.

\noindent\textbf{No graph information (w/o graph).} We omit the graph textual encoding part. Thus, the LLM is not aware of the structure of the propagation tree. Instead, only textual information about the tweets and re-tweets is provided. We observe a decrease in performance. For instance, the recall score drops from $0.8507$ to $0.795$. Similar reductions are also observed for the rest of the metrics.

\noindent\textbf{No textual information (w/o text).} In this case, we omit the textual information of the tweets of a propagation tree. We observe a significant decrease in performance. For example, the recall score drops from $0.8507$ to $0.0622$. We conclude that the textual information coupled with a propagation tree is the most important one for an LLM to detect coordinated fake news campaigns.

\begin{table}[htbp]
    \centering
    \caption{Evaluation results of the different combinations of encoded information. The best results for each metric are highlighted in bold.  The best results for each metric are highlighted in bold. Llama 3.1 70B was used in these experiments.}
    \resizebox{\columnwidth}{!}{
    \begin{tabular}{ccccc}
    \toprule
    {\bf Encoded Information} & {\bf Precision}  & {\bf Recall} & {\bf F1} & {\bf ROC AUC}  \\
    \midrule
    {\bf w/o graph} & 0.3493 & 0.7985 & 0.4860 & 0.8155  \\
    {\bf w/o text} & 0.0865 & 0.0622 & 0.0724 & 0.5059 \\
    {\bf text + graph} & \textbf {0.4238} & \textbf{0.8507} & \textbf{0.5658} & \textbf{0.8602}  \\
    \bottomrule
    \end{tabular}
    }
    \label{tab:encoded_evaluation}       
\end{table}

\section{Conclusion}
\label{sec:conclusion}

In this paper, we propose a novel framework for detecting coordinated political fake news campaigns on social media platforms like $\mathbb{X}$. Our approach leverages prompt engineering and retrieval-augmented generation, enabling LLMs to identify such campaigns by incorporating both graph-based and textual information from the news propagation tree. We also introduce `balanced RAG', a novel RAG component that considers both positive and negative examples during the retrieval step. We evaluate the effectiveness of our framework and its variants on a dataset concerning the 2016 United States presidential election. The experimental results demonstrate that the proposed framework performs better compared to traditional baseline methods, even under extreme class imbalance settings. Our framework highlights the importance of combining prompt engineering and balanced RAG with frozen LLMs to fight against organized disinformation campaigns.


\bibliographystyle{ACM-Reference-Format}
\balance
\bibliography{main}

\appendix

\section{Implementation details}
The source code is available at \url{https://github.com/nkanak/brag-fake-news-campaigns}. Due to X's privacy policies, the utilized dataset cannot be made publicly available. Instructions on how to download the dataset are provided in the GitHub repository.



\section{Discussion}
\label{sec:discussion}

\noindent \textbf{LLMs against astroturf campaigns (RQ1).} Frozen LLMs can effectively identify organized political disinformation campaigns. Our framework leverages prompt engineering and RAG and outperforms traditional GNN-based baselines. This indicates that even without fine-tuning, LLMs are able to identify coordinated disinformation efforts, given that they are provided with the appropriate prompts and relevant contextual information.

\noindent \textbf{Efficiency of text-encoded graphs (RQ2).}
Encoding graph information as text has proven to be an effective strategy for detecting astoturf disinformation campaigns. By transforming the structural data of the propagation trees into texts, we allow LLMs to analyze the relationships within the network. Utilizing graph textual encoding techniques eliminates the need for complex graph-based processing and simplifies the input while retaining critical information necessary for accurate detection.

\noindent \textbf{Utilizing only prompt engineering and RAG to identify astroturf campaigns (RQ3).}
It is feasible to utilize only prompt engineering and RAG techniques to identify astroturf campaigns using a frozen LLM. Our experiment results verify that this approach not only simplifies the detection process by avoiding the need for extensive model training but also achieves better performance compared to traditional methods.

\noindent \textbf{Prompt engineering and RAG under class imbalance (RQ4).}
By incorporating RAG examples, the framework provides the LLM with relevant instances that help balance the influence of negative samples during inference. It mitigates the bias towards real news, ensuring that the model remains sensitive to astroturf campaigns.

\noindent \textbf{GNNs for modeling graph-based data (RQ5).}
Our framework demonstrates that encoding graph information as text and utilizing LLMs can achieve better performance under class imbalance. With appropriate prompt engineering and RAG strategies, LLMs are competitive counterparts to GNNs for graph-related tasks, particularly in scenarios where training data is imbalanced.

\noindent \textbf{Importance of the proposed approach.}
The proposed framework offers several key advantages over traditional disinformation detection methods. Utilizing frozen LLMs with prompt engineering and RAG eliminates the need for labeled datasets. Additionally, the ability of the framework to effectively handle class imbalance ensures its applicability in real-world settings where astroturf campaigns are rare but important.

\section{Balanced RAG prompt example}
\label{appex:prompts}

Our proposed balanced RAG can generate pairwise examples consisting of both the positive and the negative.\newpage

\begin{lstlisting}
You are an intelligent classifier capable of identifying political fake news campaigns. Your task is to analyze the propagation of a given tweet, focusing on both the text content and aggregated network metrics to detect any coordinated behavior that might indicate a political fake news campaign. You will be provided with the tweet text and a set of statistics derived from its retweet propagation graph.

The following centrality metrics are important for your analysis:

- Degree centrality measures how many direct connections a user (node) has. A high degree centrality indicates that the user is directly connected to many others, suggesting they play a central role in propagating the tweet.
- Eigenvector centrality measures the influence of a user within the network. It considers not only the number of connections a user has, but also the importance of those connections.
- Median retweet time refers to how quickly the tweet is typically retweeted. In coordinated campaigns, retweets often occur in rapid bursts, indicating synchronized activity.

Base your decision on the text content, language use, messaging tone, and these network metrics.

Here are some labeled examples of tweets to help guide your analysis. The labels indicate whether the tweet was part of a "Fake" coordinated political campaign or "Real" organic behavior. Pay attention to the differences between Fake and Real examples.


Example 1 (Similar):
Tweet text: "[TEXT]"
Average degree centrality: 0.12
Eigenvector centrality: 0.22
Median retweet time: 4656.50 seconds
Label: Real

Example 2 (Similar):
Tweet text: "[TEXT]"
Average degree centrality: 0.07
Eigenvector centrality: 0.14
Median retweet time: 446.00 seconds
Label: Real

Example 3 (Similar):
Tweet text: "[TEXT]"
Average degree centrality: 0.07
Eigenvector centrality: 0.13
Median retweet time: 55164.00 seconds
Label: Fake

Example 4 (Contrastive):
Tweet text: "[TEXT]"
Average degree centrality: 0.13
Eigenvector centrality: 0.22
Median retweet time: 404.00 seconds
Label: Fake

Example 5 (Contrastive):
Tweet text: "[TEXT]"
Average degree centrality: 0.02
Eigenvector centrality: 0.07
Median retweet time: 3879.00 seconds
Label: Fake

Example 6 (Contrastive):
Tweet text: "[TEXT]"
Average degree centrality: 0.02
Eigenvector centrality: 0.07
Median retweet time: 44769.00 seconds
Label: Real

Now, analyze the following input tweet and determine whether it suggests a "Fake" coordinated effort or "Real" organic behavior.

Tweet text: "[TEXT]"
Average degree centrality: 0.18
Eigenvector centrality: 0.26
Median retweet time: 13865.00 seconds

Output:
Step 1: [Analyze the language and tone of the tweet]
Step 2: [Identify any sensationalism or emotional manipulation in the text]
Step 3: [Examine the network metrics and what they suggest about the tweet's spread]
Step 4: [Compare the input tweet with the provided examples, noting similarities and differences]
Step 5: [Evaluate the likelihood of coordination based on all available information]
Final Decision: [Your decision MUST be either "Fake" or "Real"]

IMPORTANT: Your final decision MUST be either "Fake" or "Real". Do not use any other labels or leave the decision ambiguous. Please replace the brackets with your actual analysis and reasoning.
\end{lstlisting}

\end{document}